\documentclass[letterpaper]{article} 
\usepackage{aaai23}  
\usepackage{times}  
\usepackage{helvet}  
\usepackage{courier}  
\usepackage[hyphens]{url}  
\usepackage{graphicx} 
\usepackage{natbib}  
\usepackage{caption} 
\usepackage{algorithm}
\usepackage{algorithmic}

\usepackage{booktabs}
\usepackage{tikz}
\usepackage{comment}
\usepackage{amsmath,amsthm,amssymb}
\usepackage{array}
\usepackage{color, colortbl}
\usepackage{bm}
\usepackage{multirow}
\usepackage{paralist}

\definecolor{darkRed}{RGB}{170,9,25}
\definecolor{darkBlue}{RGB}{24,62,170}
\definecolor{DeepBlack}{RGB}{0,0,0}

\usepackage{newfloat}
\usepackage{listings}
\DeclareCaptionStyle{ruled}{labelfont=normalfont,labelsep=colon,strut=off} 
\floatstyle{ruled}
\newfloat{listing}{tb}{lst}{}
\floatname{listing}{Listing}
\title{Truncate-Split-Contrast: A Framework for Learning from Mislabeled Videos}
\author{
    Zixiao Wang\textsuperscript{\rm 1}\thanks{This work is done during an internship at Tencent AI Lab.} \;
    Junwu Weng\textsuperscript{\rm 2} \;
    Chun Yuan\textsuperscript{\rm 1} \;
    Jue Wang\textsuperscript{\rm 2} \;
}
\affiliations{
    \textsuperscript{\rm 1} Tsinghua University \quad
    \textsuperscript{\rm 2} Tencent AI Lab \\
    {\tt\small zx-wang19@mails.tsinghua.edu.cn \quad WE0001WU@e.ntu.edu.sg \\ yuanc@sz.tsinghua.edu.cn \quad arphid@gmail.com}
}

\begin{document}

\maketitle

\begin{abstract}
Learning with noisy label (LNL) is a classic problem that has been extensively studied for image tasks, but much less for video in the literature. A straightforward migration from images to videos without considering the properties of videos, such as computational cost and redundant information, is not a sound choice. In this paper, we propose two new strategies for video analysis with noisy labels: 1) A lightweight channel selection method dubbed as Channel Truncation for feature-based label noise detection. This method selects the most discriminative channels to split clean and noisy instances in each category; 2) A novel contrastive strategy dubbed as Noise Contrastive Learning, which constructs the relationship between clean and noisy instances to regularize model training. Experiments on three well-known benchmark datasets for video classification show that our proposed tru{\bf N}cat{\bf E}-split-contr{\bf A}s{\bf T} (NEAT) significantly outperforms the existing baselines. By reducing the dimension to 10\% of it, our method achieves over 0.4 noise detection F1-score and 5\% classification accuracy improvement on Mini-Kinetics dataset under severe noise (symmetric-80\%). Thanks to Noise Contrastive Learning, the average classification accuracy improvement on Mini-Kinetics and Sth-Sth-V1 is over 1.6\%.
\end{abstract}

\section{Introduction}
Training deep networks requires large-scale datasets with high-quality human annotations. However, acquiring large-scale clean-annotated data is costly and time-consuming, forcing people to seek low-cost but imprecise alternative labeling. Such labeling inevitably introduce noises: a large number of instances could be annotated with incorrect labels. Recent studies~\cite{zhang2016understanding,arpit2017closer} have shown that deep neural networks have a high capacity to fit data even under randomly assigned labels, which harms the generalization on unseen data. Therefore, how to train a robust deep learning model in the presence of noisy labels is challenging and is of increasing significance in the industry. To date, the existing LNL approaches mainly focus on image tasks. With the rapidly growing amount of video data on the Internet, designing a noise-robust training strategy for video models becomes imperative. Motivated by the previous success of LNL methods on images, we study the much less explored problem of applying LNL methods in the video domain.


Depending on whether noisy instances are detected in training, the existing LNL methods can be roughly divided into two types. One is to directly train a noise-robust model in the presence of noisy labels~\cite{patrini2017making,wang2019symmetric,ma2020normalized,lyu2019curriculum,zhou2021learning,gao2021searching}. The other one is to explicitly detect the potential noisy instances, and then learns a model by simply excluding them~\cite{huang2019o2u}, or re-using the potential noisy data by estimating the pseudo labels of them~\cite{zhang2018boosting,li2019dividemix,li2021learning,ortego2021multi}. This detection strategy is widely adopted in the industry as it not only learns a robust model, but provides a clean dataset as well. Following this strategy, we learn video representations from potentially-mislabeled data with two steps, Noise Detection and Unlabelled Data Utilization.

\noindent\textbf{Noise Detection.} The \textit{loss-based} method~\cite{han2018co,zhang2018boosting,huang2019o2u,li2019dividemix} is the common solution for noisy label detection. These methods treat instances with smaller classification losses as clean ones during training, which easily leads to confirmation bias~\cite{li2019dividemix}. Compared with a single loss value, video latent representations naturally contains multi-channel signals, which provide adequate clues for noise detection. 

The existing {\it feature-based} noisy label detection methods~\cite{lee2018cleannet,han2019deep} commonly conduct binary clustering (clean/noisy) on the full-dimensional features before the classification layer. However, we argue that to detect clean/noisy instances, utilizing all channels of a feature learned from classification supervision is not a must. The reason is that the instance feature learned with label supervision may perform well on differentiating categories, but the extra feature dimensions for delicate boundary shaping are not suitable for an unsupervised binary clustering task in label noise detection, especially under the video domain. We find that the extra dimensions are redundant and weaken the performance in noise detection and therefore in final classification. (Sec.~\ref{sec:ablation} - Table.~\ref{tab:b}) Designing a compact network may help reduce the redundancy of feature channels for noise detection, but the limited network capacity will inevitably hurt the classification, as these two tasks share the same network and learned features. Channel selection therefore becomes our first choice.

In this paper, we propose a {\it category-wise} channel selection method, {\it i.e.} Channel Truncation (CT), for feature-based label noise detection in videos. It evaluates the discriminative ability of each channel of instance representations by simply collecting the temporal statistics across frames of each instance.
After sorting all channels based on their noise-discriminative abilities, CT adaptively removes the most confusing ones for each category during training. Afterward, each truncated instance feature is matched with its relevant category clean-prototype to determine whether it is clean or not. This CT for videos can also be simplified to detect noisy label of image data (Appendix).

\noindent \textbf{Unlabelled Data Utilization.} The previous LNL methods commonly process the detected noise under a semi-supervised learning framework. A pseudo label is assigned to each noisy instance to replace the wrongly annotated one as a supervision signal for model training~\cite{li2019dividemix,li2021learning,ortego2021multi}. One drawback of pseudo labeling is that the pseudo label could be ambiguous and unreliable without sophisticated post-processing and data enrichment~\cite{zhang2018mixup}. This phenomenon is even severe under video domain as the redundant information in videos may mislead the pseudo labeling at the early training stage. In the experiment, we verify that the naive pseudo label has negative impact to the video classification (Sec.~\ref{subsec:ncl} - Table.~\ref{tab:k200}), while the impact on image is not that severe. Therefore, a new design of unlabelled data utilization in videos is needed. On the other side, little effort has been directed towards enhancing the quality of instance representation when no label is assigned to the detected noise. Compared with guessing the actual label of the noise, the mutual relationship among the instances is relatively easy to estimate after the clean/noise splitting. Inspired by contrastive learning~\cite{oord2018representation,jing2020self}, we propose a \textit{Noise Contrastive Loss} (NCL) to utilize the unlabelled noise and enlarge the margins among instances from different categories. NCL provides a low-risk contrastive strategy for unlabelled noisy queries, avoiding misguidance from wrong pseudo labels. 

The framework of tru{\bf N}cat{\bf E}-split-contr{\bf A}s{\bf T} ({\bf NEAT}) is shown in Fig.~\ref{fig:pipeline}. Each video feature is first {\it truncated} for clean/noisy instance {\it splitting}. Then, the detected clean/noisy instances are utilized separately under the supervision of cross entropy and noise {\it contastive} loss for model updating. Our main contributions are summarized as follows: 
\begin{compactitem}
    \item A lightweight channel selection method for feature-based label noise detection is proposed. It discards the redundant channels to increase the effectiveness and efficiency of noisy/clean instance splitting.
    \item Noise Contrastive Loss is designed to construct the relationship among instances by referring the estimated clean/noisy splits, and utilizes this relationship to learn visual representations without involving wrong labels.
    \item To the best of our knowledge, this is the first efficient framework for LNL in video analysis. Extensive experiments show the effectiveness of our method on several video recognition datasets with noisy label settings.
\end{compactitem}

\section{Related Work}


\subsection{Learning with Noisy Label}
\label{Sec 2.1}
LNL on images has been extensively studied in the literature. In this section, we only limit our review to the noise detection methods and the way these methods utilize the detected noisy instances. 

\noindent\textbf{Noise Detection.} There are two main streams of noise detection methods, the loss-based one and the feature-based one. We here mainly discuss feature-based methods. Feature-based methods \cite{kim2021continual} detect noisy samples based on feature similarities between samples. \cite{lee2018cleannet} takes the cosine similarity between unidentified samples and the prototypes as references to detect noise. \cite{wu2020topological} utilizes $k$-nearest neighbor ($k$-NN) to build a neighbor graph for each category, treating samples in the dominant sub-component as clean ones. \cite{ortego2021multi} also uses $k$-NN and voting to determine whether a sample is clean or not. These works mainly focus on dealing with label noise in image tasks, and so far, no work discusses how things are different in video scenarios. Besides, they refer to the full-dimensional feature for noise detection, which we argue is unnecessary in this task.
\begin{figure*}[t] 
	\centering
		\includegraphics[width=1.0\linewidth]{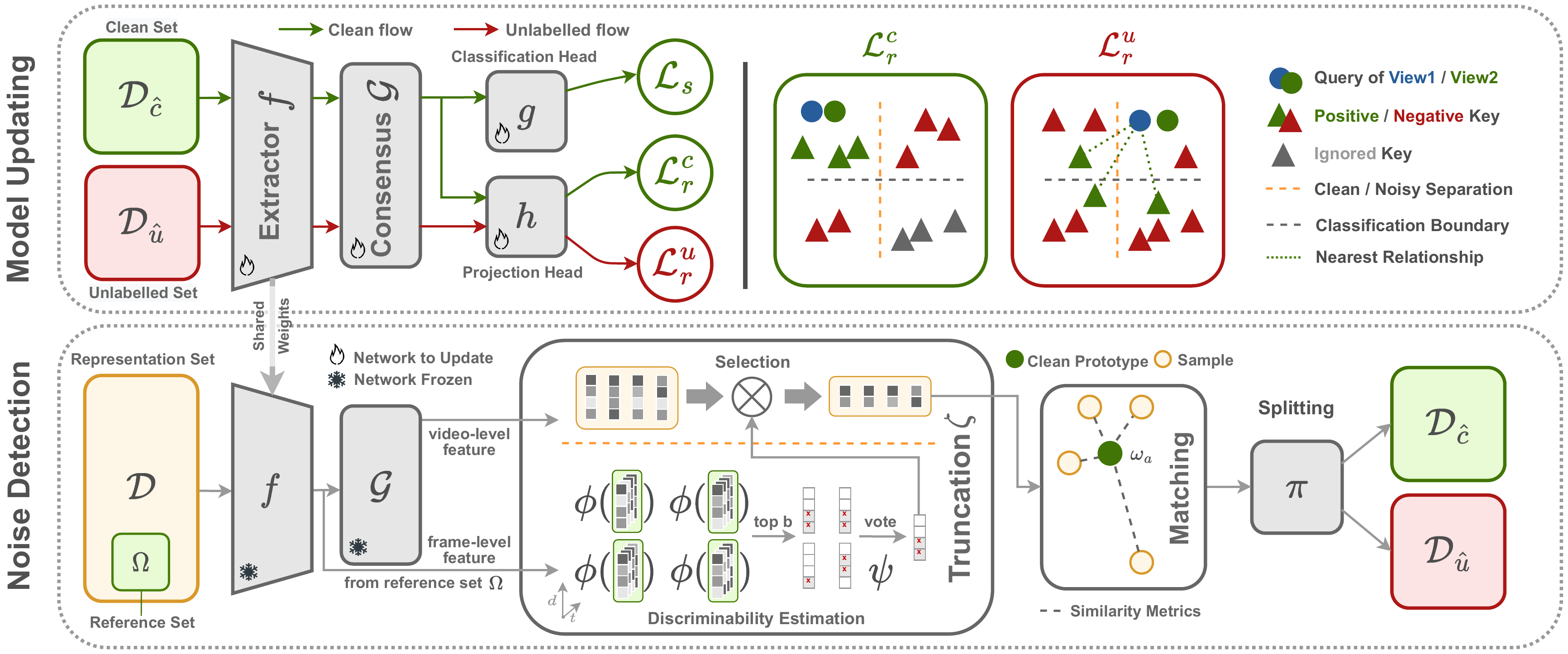}
	
	\caption{The pipeline of our framework NEAT on the noisy dataset in training. There are two phases in the framework. In the {\it Noise Detection} phase, the whole training dataset is split-ed into clean/noise clusters by considering the similarities between each pair of dimension-reduced instance representations. In this phase the network is frozen for feature extraction only. During {\it Model Updating} phase, the detected clean instances are fed into the network for supervised learning, while all instances are utilized in Noise Contrastive Learning for decision boundary shaping. These two phases proceed iteratively. In the first round of model updating, all the instances are utilized for supervised learning.}
	\label{fig:pipeline}
\end{figure*}

\noindent\textbf{Noise Utilization.} The common strategy for noise utilization is to assign a pseudo label to the detected noisy instance as a supervision label. DivideMix~\cite{li2019dividemix} estimates the pseudo label from model prediction and applies Mixup~\cite{zhang2018mixup} to enhance the reliability of the pseudo label. \cite{ortego2021multi} combines MixUp~\cite{zhang2018mixup} and Supervised Contrastive Learning~\cite{khosla2020supervised} to mitigate the negative impact of noisy labels in representation learning. The label correction in this work is achieved by $k$-NN search. Junnan Li {\it et.al} improve prototypical contrastive learning (PCL)~\cite{li2020prototypical} in~\cite{li2021learning} by using pseudo-labels to compute class prototypes. This work applies a similar $k$-NN-based label correction strategy as~\cite{ortego2021multi}. The drawback of pseudo labeling is that the corrected label is sometimes not reliable, and label enhancement techniques like sharpening~\cite{li2019dividemix} are needed. Compared with the pseudo labeling, instead of estimating the unreliable pseudo labels or involving the noisy labels, our proposed NCL utilizes the mutual relationship between the distribution of clean and noisy instances to shape the classification boundary such that the negative impact of estimated labels is mitigated. No complex post-processing and data enrichment is required. 

\subsection{Contrastive Representation Learning}
\label{Sec 2.2}
In recent years, the contrastive representation learning methods dominant the existing literature in self-supervised learning~\cite{oord2018representation,jing2020self}. They optimize similarities of positive (negative) pairs to improve the quality of representations. Techniques like data augmentation~\cite{tian2020makes}, larger batch size~\cite{chen2020simple,khosla2020supervised}, network design~\cite{chen2020improved} are always used for better representation learning. Regarding the label noise scenario, the Instance Contrastive Learning~\cite{oord2018representation}, the Supervised Contrastive Learning~\cite{khosla2020supervised} and the Prototypical Contrastive Learning (PCL)~\cite{li2020prototypical} are modified in~\cite{kim2021continual}, ~\cite{ortego2021multi} and~\cite{li2021learning} correspondingly and respectively to fully involve the detected noisy instances.

\section{Method}


Under the video scenario, $T$ frames are sampled from each video as a clip. The normalized feature of a frame is defined as $\bm{v}$, which is extracted from a backbone network $f(\cdot)$. Afterward, the representation of a clip from a video is defined as $ \bm{x}= \mathcal{G}\left(\bm{v}_1, \dots, \bm{v}_{T}\right)$, where $\mathcal{G}$ is a consensus function combining the set $\{\bm{v}_t\}^T_{t=1}$ with normalization. Given a representation dataset $\mathcal{D}=\{\bm{x}_i\}^M_{i=1}$ with $M$ normalized $\bm{x} \in \mathbb{R}^d$, the goal of a classification task is to find which category $\bm{x}$ belongs to. The annotation of $\bm{x}$ is $a$, and it can be also represented as an one-hot vector $\bm{y} \in \{0,1\}^K$, in which the $a$-th element of $\bm{y}$ is assigned as $1$, and the remainings are assigned as $0$s. Here $K$ indicates the number of category. A classification head $g(\cdot)$ with Softmax operation is defined to predict the probability of $\bm{x}$ belonging to the $k$-th category, namely $p(k|\bm{x}) = g(\bm{x},k)$.

With the noisy label existing, a label $a$ may be wrongly assigned to a sample not belonging to the $a$-th category set. To reduce the negative impact of wrongly annotated samples in model training, our strategy is to first estimate which samples are correctly annotated, {\it i.e.} clean samples from $\mathcal{D}_c$, and which are not, {\it i.e.} noisy samples from $\mathcal{D}_u$, where $\mathcal{D} = \mathcal{D}_c \cup \mathcal{D}_u$. The noisy samples are regarded as unlabelled in our framework. This noise detection phase is achieved by a detection function $\mathcal{D}_{\hat{c}} = \epsilon(\mathcal{D})$. In the following model updating phase, the labels of the estimated clean samples are directly used in cross entropy loss for label supervision. Meanwhile, the detected noisy samples are involved in model updating as regularization where the impact of wrong labels is ignored. With the notations described above, the model updating process under the noisy label setting can be guided by the loss defined as below,
\begin{equation}
\begin{aligned} 
    \mathcal{L} &= \mathcal{L}_{s} + \mathcal{L}_r  \\
                &= -\sum_{\bm{x}_i\in \epsilon(\mathcal{D})} \sum^{K}_{k=1} \bm{y}_i(k)\cdot\log\,g(\bm{x}_i,k) + \mathcal{L}_r,
\end{aligned}\label{eq:ce} 
\end{equation}
in which the loss $\mathcal{L}_{s}$ is defined by cross entropy for label supervision. The detection function $\epsilon(\cdot)$ filters out the noisy samples in the training set for each category by similarity measure, in which a dimensionality reduction method Channel Truncation, is proposed for the discriminative channel selection. The loss $\mathcal{L}_r$ is designed as a regularization in model updating involving both estimated clean and noisy samples. We introduce Noise Contrastive Learning in $\mathcal{L}_r$ to utilize the relationships among clean and noisy instances for decision boundary shaping. The pipeline of our framework is shown in Fig.~\ref{fig:pipeline}.

\subsection{Channel Truncation}

Generally, feature-based noise detection methods detect noisy labels by computing the similarity between the query $\bm{x}$ and the clean-prototype $\bm{x}_a$ of category $a$. The higher the similarity, the more likely the query $\bm{x}$ to be clean. As full channels are learned to differentiate multiple categories, they are unnecessary for a much simpler task, {\it i.e.} differentiating clean/noisy instances in each category. Truncating inessential channels can help detect clean instances better than utilizing all. Therefore, we propose Channel Truncation to truncate the redundant channels and keep the discriminative ones for clean instance detection. 

Our method owns a top $b$ channel selection operation ${\zeta}_b(\cdot)$ with a category-level score function $\psi(\cdot)$. ${\psi}(\cdot)$ evaluates the discriminative ability of each channel in a category, such as $a$, by referring to the statistics of its related reference set ${\Omega}^a$. This score function returns a vector with the same dimension as feature $\bm{x}$, where each element is a score of the corresponding feature channel. Selection operation ${\zeta}_b(\cdot)$ picks $b$ channels with the highest scores of the input representation and returns a dimension-reduced $\bm{w}$ concatenating these top channels. Therefore the truncation function is defined as,
\begin{equation}
     \bm{w} = \zeta_b\big(~\bm{x} ,~\psi\,({\Omega}^a)\big). \label{eq:ct}
\end{equation}

Given truncated-feature $\bm{w}$ and its label $a$, the similarity measure between $\bm{w}$ and the class clean-prototype $\bm{w}_a$ is obtained by a similarity function ${\eta}_a \left(\cdot\right)$, which is defined as inner product, namely ${\eta}_a \left(\bm{w}\right) = \bm{w}\cdot \bm{w}_a$. This similarity indicates how close the instance $\bm{w}$ to the clean cluster of category $a$. The class clean-prototype $\bm{w}_a$ is defined as the average $\bm{w}$ of the estimated clean set $\mathcal{D}_{\hat{c}}$. We observe that the similarity distribution of the clean and noisy instances gradually becomes a two-peak form during training. Thus to detect the noisy videos, a two-component Gaussian Mixture Model $\pi(\cdot)$ is utilized to fit the distribution of ${\eta}(\bm{w})$. The probability of $\bm{w}$ being noise of the $a$-th category is then defined as $p({\rm noise}|\bm{w} ; a) = \pi_{a}(\bm{w})$. Hence, the estimated clean set is obtained by thresholding $\pi_{a}(\bm{w})$,
\begin{equation}
    \mathcal{D}_{\hat{c}} = \epsilon(\mathcal{D}) = \{\bm{x}~|~\bm{x} \in \mathcal{D},~\pi_{a}(\bm{w}) < \xi\}. \label{eq:select}
\end{equation}
where $\xi$ is a threshold and we fix it as $0.5$ in the experiments. For simplicity below, we ignore the category index $a$. Next, how to design a proper score function ${\psi}(\cdot)$?

\noindent\textbf{Oracle Selection} Ideally, when the true splits of correctly and wrongly annotated instances are known beforehand, we take all the clean training data from the category $a$ as reference set $\Omega$, and the channel discriminative ability can be measured by the within-/~between-cluster variance. Following the Fisher Discriminant Analysis, for a specific category $a$, the oracle  score function ${\psi}_o(\cdot)$ is then defined as:
\begin{equation}
\begin{aligned}
    &\psi_o = \frac{\left(\bm{\mu}_c-\bm{\mu}_u\right)^2}{\bm{\sigma}_c^2 + \bm{\sigma}_u^2}  \\ 
    &\bm{\mu}_c = \frac{1}{|\Omega|}\sum_{\bm{x}_i\in \Omega}\bm{x}_i,~~\bm{\sigma}_c^2 = \frac{1}{|\Omega|}\sum_{\bm{x}_i\in \Omega}\big(\bm{x}_i-\bm{\mu}_c\big)^2, \label{eq:score_lda}
\end{aligned} 
\end{equation}
where $\bm{\mu}_c \in \mathbb{R}^d$ and $\bm{\sigma}_c \in \mathbb{R}^d$ are the average values and standard deviation of all $\bm{x}_i$ in set $\Omega$, respectively. All the operations defined here in the equation are element-wise. Similarly, $\bm{\mu}_u$ and $\bm{\sigma}_u$ are the statistics of the unlabelled (mislabeled) set. The higher the score, the more discriminative the corresponding channels. As the true splits of correctly and wrongly annotated instances are unknown in training, the oracle score in Eq.\ref{eq:score_lda} {\it cannot} be used as the selection criterion of channels. However, this measurement can be treated as an evaluation metric of practicable ${\psi}(\cdot)$.

\begin{figure}[!tb]
    \centering
    \includegraphics[width=1\linewidth]{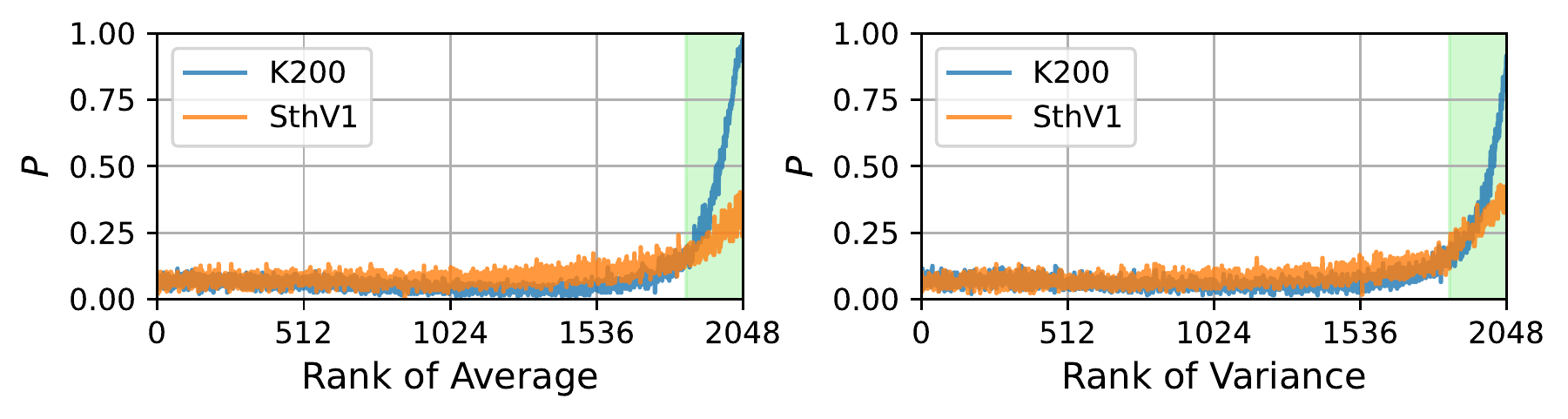}
    \caption{The statistical relation between oracle score and amplitudes/variance on K200 and SthV1 under symmetric-40\% noise setting at fifth epoch. Each recorded point bears the coordinates $(r,p)$. $r$ is the ranking of the corresponding statistics. The higher the $r$, the larger the amplitudes/variance. $p$ is the probability of the relevant channels picked by the top $b$ {\it oracle selection}. The top $b$ amplitudes/variance area is filled with green. ($b=200$)}
    \label{fig:lda}
\end{figure}

\noindent\textbf{Proposed Selection} A video, in essence, consists of both the scene and motion semantics~\cite{wang2018pulling,choi2019can,weng2020temporal,wang2021removing}. The information of global scene appearance generally remains almost unchanged in a video, while the motion-relevant information varies on the temporal domain. To well differentiate videos, it is critical to distill both signals. We therefore utilize the temporal {\it average} and {\it variance} to roughly search the semantics-intense channels.
As expected, we experimentally discover that channels with larger temporal {\it average} and {\it variance} of amplitudes tend to achieve higher oracle scores, as shown in Fig.~\ref{fig:lda}.
Thus, we propose to estimate the channel discriminative ability by introducing an instance-level score function $\phi(\cdot)$ and defining category-level score function $\psi(\cdot)$ as a histogram counting the top $b$ instance-level score events of each channel in a certain category. The function $\phi(\cdot)$ is designed as temporal statistics of $\{\bm{v}_t\}_{t=1}^T$ from $G$, which is lightweight and suitable for the video scenario. Here we have two versions of $\phi(\cdot)$, the average pooling $\phi_{ave}$, and the temporal variation $\phi_{var}$ of the representation channel. They are defined as, 
\begin{equation}
\begin{aligned}
    &\phi_{ave}(\bm{v}_1, \dots, \bm{v}_{T}) = {\sum}_t \bm{v}_t/T, \\
    &\phi_{var}(\bm{v}_1, \dots, \bm{v}_{T}) = {\sum}_t (\bm{v}_t - \phi_{ave})^2/T.
\end{aligned}
\end{equation}

We set $\Omega$ as the clean set $\mathcal{D}_{\hat{c}}$ from the last epoch and initialize $\Omega$ as the dataset $\mathcal{D}$ in the first training epoch. The {\it average} responses extract the similar signal among $T$ frames. The {\it variance} of the amplitude distills the dissimilar signal across the frames. Each operation here is element-wise. It is observed that the selected channels are closely related to the semantics, {\it e.g} scene and motion, of a video clip in the specific category, about which we will show more visualizations in the Appendix. Moreover, a further detailed analysis of different temporal statistics will be discussed in Sec.\ref{sec:experiment}.

\subsection{Noise Contrastive Learning}
\label{subsec:ncl}
With the estimated clean and unlabelled splits, namely $\mathcal{D}_{\hat{c}}$ and $\mathcal{D}_{\hat{u}}$, the motivation of Noise Contrastive Learning (NCL) is to fully utilize the estimated noisy samples in model updating, and further enlarge the margins among samples from different categories. 

To reach this goal, contrastive learning~\cite{oord2018representation} is involved in NCL to enforce the consistency within each clean cluster and enlarge the dissimilarity between clean and noisy clusters. We first randomly sample two clips from a video to represent two different views of this video. Both view clips consist of $T$ frames following a certain sampling strategy. In NCL, we take one clip representation of a video as a query and set the remaining clips from the same and other videos as keys. The selection of positive and negative keys varies from clean query to noisy query. For a query representation $\bm{x}_i \in \mathcal{D}_{\hat{c}}$ from {\it clean} cluster of category $a_i$, its sets of positive and negative keys $\mathcal{P}^{i}_{\hat{c}}$, $\mathcal{N}^{i}_{\hat{c}}$ can be defined as,
\begin{equation}
\begin{aligned}
    \mathcal{P}^{i}_{\hat{c}}=&\{\bm{x}_j|\bm{x}_j \in \mathcal{D}_{\hat{c}}, i \small{\neq} j, a_i \small{=} a_j\},  \\
    \mathcal{N}^{i}_{\hat{c}}=&\{\bm{x}_j|\bm{x}_j\in \mathcal{D}_{\hat{u}}, a_i \small{=} a_j\}\cup\{\bm{x}_l|\bm{x}_l \in \mathcal{D}_{\hat{c}}, a_i \small{\neq} a_l\},
\end{aligned}
\end{equation}
respectively. By this way, the estimated clean instances in the same category are forced to be close to each other, and the ones from different categories are pushed away from one another. When the query representation $\bm{x}_i \in \mathcal{D}_{\hat{u}}$ is from {\it unlabelled} cluster, its sets of keys $\mathcal{P}^{i}_{\hat{u}}$ and $\mathcal{N}^{i}_{\hat{u}}$ are defined as,
\begin{equation} 	
\begin{aligned}
 	\mathcal{P}^i_{\hat{u}}=&\{\bm{x}_j|\bm{x}_j \in {\rm NN}(\bm{x}_i)\} \cup\{\bm{\widetilde{x}}_i\},\\
 	\mathcal{N}^i_{\hat{u}}=&~\mathcal{D}-\mathcal{P}_{\hat{u}}\cup\{\bm{x}_i\},
\end{aligned}
\end{equation}
in which $\bm{x}_i$ and $\bm{\widetilde{x}}_i$ are the two views of a query video. The positive keys of the noisy query also come from the nearest neighbours. Specifically, we retrieve the top similar keys of a query $\bm{x}_i$ by $k$-NN function ${\rm NN}(\bm{x})$ which returns the $B$ nearest neighbors of input instance in the set $\mathcal{D}-\{\bm{x}\}$. With the proposed positive and negative sets $\mathcal{P}_{\hat{c}}$, $\mathcal{P}_{\hat{u}}$, $\mathcal{N}_{\hat{c}}$ and $\mathcal{N}_{\hat{u}}$, we extend the InfoNCE~\cite{oord2018representation} loss to
two noisy contrastive losses $\mathcal{L}^c_r$ and $\mathcal{L}^u_r$ to force the clean and noisy clusters to be apart from each other:
\begin{equation}
\begin{aligned}
\label{eq:ncl}
  \mathcal{L}_r =& \frac{1}{|\mathcal{D}_{\hat{c}}|} \sum\limits_{\bm{x}_q \in \mathcal{D}_{\hat{c}}}\mathcal{L}^c_r(\bm{x}_q) + \frac{1}{|\mathcal{D}_{\hat{u}}|} \sum\limits_{\bm{x}_q \in \mathcal{D}_{\hat{u}}}\mathcal{L}^u_r(\bm{x}_q)
\end{aligned}
\end{equation}
where $\mathcal{L}^c_r=\gamma(\mathcal{P}_{\hat{c}}, \mathcal{N}_{\hat{c}})$ and $\mathcal{L}^u_r=\gamma(\mathcal{P}_{\hat{u}}, \mathcal{N}_{\hat{u}})$. $\gamma(\cdot)$ is defined as:
\begin{equation}
\label{eq:ncl_gamma}
  \gamma(\mathcal{P}, \mathcal{N}) = -\frac{1}{|\mathcal{P}|}\sum\limits_{\bm{x}_{+} \in \mathcal{P}} \log{\frac{\exp{\left(\bm{z}_q\cdotp \bm{z}_{+}/{\tau}\right)}}{\sum\limits_{\bm{x}_j\in \mathcal{P}\small{\cup} \mathcal{N}}{\exp{\left(\bm{z}_q\cdotp\bm{z}_j/{\tau}\right)}}}},
\end{equation}
in which $\bm{z}\in\mathbb{R}^{\hat{d}}$ is the normalized representation mapped by projection head $h(\cdot)$, {\it i.e.} $\bm{z} = h(\bm{x})$, and $\hat{d}<d$. The symbol ``$\cdotp$" denotes the inner (dot) product between two vectors. $\tau$ is the temperature scaling.


Compared with assigning pseudo labels to the estimated unlabelled instances~\cite{li2019dividemix}, the proposed NCL is low risk as it does not involve the label of the estimated noisy samples, which may severely mislead the model updating. Even though the true labels of the noisy samples are ignored in label supervision of the model updating, the {\it mutual exclusion} between the clean and noisy instances in each category can still well shape the decision boundary indirectly. In each category, most of the detected noisy instances do not share the same label as the clean ones. Therefore when expanding the gap between clean and noisy clusters, the decision margin between each category pair is simultaneously enlarged. The NCL and CT nourish each other. The model regularized by NCL provides a good representation for CT to differentiate clean and noisy instances better. Meanwhile, a well splitting of clean / unlabelled instances by CT supplies reasonable separation of positive and negative key sets for Noise Contrastive Learning.

\section{Experiments}
\label{sec:experiment}

%

\subsection{Settings and Datasets}

\noindent{\bf Model} 
In this section, We choose TSM-ResNet50~\cite{lin2019tsm} with ImageNet-pretraining as our backbone for all experiment settings. All training hyper-parameters are adapted from the original work~\cite{lin2019tsm} and detailed in Appendix. Following previous work, we choose $\tau=0.1$ for all experiments in this paper. A memory bank with the size of $16\times K$ is added for all contrastive learning loss to provide 
ample training positive-negative pairs. All baselines in this paper are re-implemented by open-source codes.

\begin{table}[t]
    \caption{Testing Accuracies (\%) on Mini-Kinetics Dataset.}
    \label{tab:k200}
	\centering
    \small
    \setlength{\tabcolsep}{3.2pt}
		\begin{tabular}{l|cccc|ccc}
			\toprule
			\multicolumn{1}{c|}{Noise Type}&\multicolumn{4}{c}{Symmetric}&\multicolumn{3}{c}{Asymmetric} \\
			\multicolumn{1}{c|}{Noise Ratio}&20\%&40\%&60\%&80\%&10\%&20\%&40\%\\

			\midrule

            GCE\shortcite{zhang2018generalized}&53.1&49.6&42.1&23.4&54.0&52.0&41.4\\
            SCE\shortcite{wang2019symmetric}&64.5&57.8&48.1&27.9&67.2&62.0&46.9 \\
            TopoFilter\shortcite{wu2020topological}&61.4&55.5&37.7&14.9&65.7&63.6&55.5 \\
            Co-teaching\shortcite{han2018co}&61.0&60.9&56.9&32.5&60.6&60.2&46.9\\
            M-correction\shortcite{arazo2019unsupervised}&66.7&62.3&54.8&40.1&65.5&62.1&52.9\\
            CT-{\it all} &68.4&64.2&58.3&43.3&69.1&67.4&51.9 \\
            CT-{\it var}&69.2&\textbf{67.1}&\textbf{61.1}&\textbf{48.4}&\textbf{70.0}&68.0&55.9\\
            CT-{\it ave}&\textbf{69.4}&66.9&61.0&48.1&69.8&\textbf{68.6}&\textbf{58.4}\\
            
            \rowcolor{green!10} CT-{\it oracle} &70.4&67.7&61.5&49.6&70.6&70.0&58.9 \\
            \rowcolor[gray]{0.85} Clean Only &70.5&68.3&64.9&58.8&70.9&70.3&68.6 \\
            \midrule
            
            DivideMix\shortcite{li2019dividemix}&69.4&65.9&60.7&46.5&68.2&67.5&53.1\\
            CT-{\it var} + PL-H&68.1&65.6&58.3&37.9&66.5&65.8&54.6\\
            CT-{\it var} + PL-S&68.2&66.0&60.8&45.4&67.1&66.8&55.8\\
            CT-{\it var} + PL-K&67.7&65.0&60.1&47.4&66.4&65.7&55.0\\
            CT-{\it var} + CL&69.4&66.6&60.4&12.2&68.3&67.6&54.7\\
            CT-{\it var} + SCL&70.3&67.6&61.4&46.9&70.2&68.1&57.0 \\
            CT-{\it var} + NCL&\textbf{70.9}&\textbf{68.6}&\textbf{63.4}&\textbf{49.9}&\textbf{70.5}&\textbf{69.5}&\textbf{59.2}	 \\

			\bottomrule 
			
		\end{tabular}
	
\end{table}

\begin{table*}[t]
    \caption{Testing Accuracies (\%) on Kinetics and Something V1 Dataset.}
    \label{tab:k400}
	\centering
    \small
    \setlength{\tabcolsep}{3pt}
		\begin{tabular}{l|ccc|ccc|ccc|ccc}
			\toprule
			\multicolumn{1}{c|}{Dataset}&\multicolumn{6}{c|}{Kinetics}&\multicolumn{6}{c}{Something V1} \\
			\multicolumn{1}{c|}{Noise Type}&\multicolumn{3}{c|}{Symmetric}&\multicolumn{3}{c|}{Asymmetric}&\multicolumn{3}{c|}{Symmetric}&\multicolumn{3}{c}{Asymmetric} \\
 			\multicolumn{1}{c|}{Noise Ratio}&40\%&60\%&80\%&10\%&20\%&40\%&40\%&60\%&80\%&10\%&20\%&40\%\\

			\midrule
            TopoFilter\shortcite{wu2020topological}&49.1&40.4&21.5&56.1&55.1&47.8&21.4&4.3&1.2&35.8&34.5&26.9\\
            Co-teaching\shortcite{han2018co}&53.9&51.4&31.3&51.4&51.4&41.7&23.6&14.5&4.5&24.0&24.6&18.5\\
            M-correction\shortcite{arazo2019unsupervised}&57.3&51.2&40.3&54.8&53.9&47.1&24.1&15.1&4.5&36.6&35.2&22.0\\
            CT-{\it ave}&\pmb{60.3}&\pmb{56.6}&\pmb{46.3}&\pmb{62.9}&\pmb{61.3}&48.0
            &\pmb{36.3}&26.2&4.6&41.5&37.8&28.3\\
            CT-{\it var}&60.1&55.8&45.7&62.8&61.0&\pmb{48.4}&36.2&\pmb{26.7}&\pmb{4.8}&\pmb{41.6}&\pmb{38.7}&\pmb{28.7}\\
            \rowcolor[gray]{0.85} Clean Only&61.6&58.8&54.3&70.3&68.0&61.7&40.7&36.0&24.8&44.0&43.5&40.6\\
            \midrule
            
            CT*-{\it var}&61.1&56.7&48.6&63.6&62.5&57.8&36.8&27.0&5.8&41.7&40.2&32.8\\
            \midrule
            CT-{\it var}+CL&60.0&55.3&7.6&61.7&59.5&45.2&31.8&1.6&1.1&37.8&36.2&27.1\\
            CT-{\it var}+SCL&60.5&56.5&46.5&63.0&60.9&48.8&37.6&28.4&2.9&41.5&40.2&29.4\\
            CT-{\it var}+NCL &\pmb{61.2}&\pmb{57.2}&\pmb{46.9}&\pmb{63.3}&\pmb{61.5}&\pmb{49.1}&\pmb{38.3}&\pmb{30.1}&\pmb{5.1}&\pmb{41.8}&\pmb{40.8}&\pmb{30.0}\\

			\bottomrule 
			
		\end{tabular}
	
\end{table*}

\begin{figure*}[t]
\centering

\includegraphics[width=0.75\linewidth]{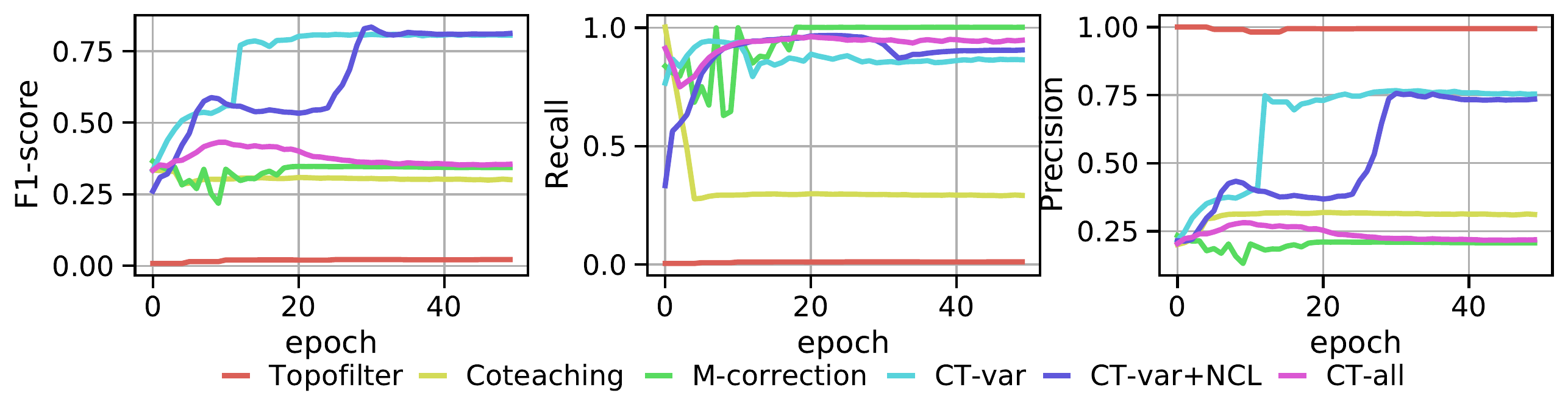}

\caption{F1-Score/Precision/Recall in K200 dataset with 80\% symmetric label noise.}
\label{fig:F1}
\end{figure*}

\noindent{\bf Datasets} 
We conduct experiments on three large-scale video classification datasets, Kinetics-400 (K400), Mini-kinetics (K200), and Something-Something-V1 (SthV1). 

\noindent$\bullet$ \textbf{K400} This is a large-scale action recognition dataset with 400 categories, and it contains $\sim$240k videos for training, and $\sim$20k (50 per category) for validation. Note that K400 is not a balanced dataset. The amount of training samples in each category are in the range of $\left[250,1000\right]$.

\noindent$\bullet$ \textbf{K200} This is a balanced subset of K400 with 200 categories. In this dataset, each category contains 400 videos for training and 25 videos for validation. 

\noindent$\bullet$ \textbf{SthV1} Something-Something is a challenging dataset focusing on temporal reasoning. In this dataset, the scenes and objects in each category are various, which strongly requires the classification model to focus on the {\it temporal dynamics} among video frames. SthV1 is an unbalanced dataset with 174 categories. It contains $\sim$86k videos for training and $\sim$12k for validation. 

\noindent{\bf Noise Types}
There are two kinds of label noise in the experiment: symmetric and asymmetric noise. Hyper-parameter $\rho$ controls the noise ratio.

\noindent$\bullet$~\textbf{Symmetric Noise}: Each noisy sample in the training set is independently and uniformly assigned to a random label other than its true label.

\noindent$\bullet$\textbf{Asymmetric Noise}: The samples in one category can only be assigned to a specific category other than the true label.

\subsection{Label Noise Detection}

We compare the Channel Truncation (CT) with multiple well-known methods for LNL. GCE~\cite{zhang2018generalized} and SCE~\cite{wang2019symmetric} are robust loss functions. We also involve Clean Only baseline in the experiment as reference, which is trained with the correctly annotated samples. TopoFilter~\cite{wu2020topological} is a feature-based method that detects noise by an L2-norm-based clustering. Co-teaching~\cite{han2018co} is a loss-based method with a dual-network design. The noise ratio $\rho$ is assumed to be already known in Co-teaching. M-correction~\cite{arazo2019unsupervised} is also a loss-based method with beta mixture model for clean/noisy instance separation. DivideMix\cite{li2019dividemix} applies Mix-up\cite{zhang2018mixup} on both labeled and unlabeled samples to enrich the dataset. As they are designed for images, we here re-implement video versions of them for comparison.

All methods share the same network architecture (TSM-ResNet50), training hyper-parameters, noise settings, and train/test splits of datasets for a fair comparison. We set $b$ to $200$ in all settings. Since each query is excepted to have 16 true positive keys in the memory bank with size $16\times K$, the number of nearest keys $B$ is set as $16$. The reference set $\Omega$ is initialized with the whole dataset $\mathcal{D}$ except for Asymmetric-40\%. In the setting of Asymmetric-40\%, we use $k$-means for binary clustering and regard the cluster with the dominant instances as the initialization of $\Omega$. Note that we choose $\bm{\phi}_{var}$ as the score function for all experiment and denote our method with $\bm{\phi}_{var}$ as CT-{\it var}. As an additional comparison, CT-{\it all} takes all the channels for noise detection, {\it i.e.} $b=2048$. As a tight upper bound of CT, CT-{\it oracle} is implemented by utilizing oracle selection for channel selection.

Table.~\ref{tab:k200} and \ref{tab:k400} shows the best testing accuracy across all epochs. The best results are marked as \textbf{bold}. On the K200, K400, and SthV1 datasets, our method greatly outperforms other LNL baselines under different levels of symmetric/asymmetric label noise. 

Due to the unbalanced number of samples in different categories, under the asymmetric noise with large $\rho$, some categories' number of noisy instances may be larger than that of clean ones. The clean/noisy instances in these unbalanced datasets are theoretically indistinguishable if extra information is not provided. To tackle this issue, we keep a balanced instance set, where each category contains {\it ten} samples with correct labels. By taking this small clean sets as $\Omega$, CT* achieves a substantial improvement in these unbalanced situations. 

Fig.~\ref{fig:F1} shows the F1-score, Precision, and Recall of noise detection during training on K200 with symmetric-80\% noise. Our method beats the baselines in almost every training epoch in F1-Score. We notice that most methods have a similar F1-score at the very beginning, and no one has an obvious advantage over others when the model is weak. It takes about ten epochs for CT-{\it var} to warm up before noise detection ability leaps. The warm-up duration of CT-{\it var}+NCL is much larger than that of CT-{\it var}, which may be caused by the randomly-initialized projection head $h(\cdot)$. Thanks to NCL, our method tends to discover more clean instances leading to higher recall but a little bit lower precision than CT-{\it var}. The improved classification accuracy shows that this Recall-Precision trade-off is beneficial.

\begin{table}[t]
    \caption{Testing Accuracies (\%) on K200 with 60\% Symmetric Noise and Different Number of Kept Channels $b$.}

	\centering
    \small
    \setlength{\tabcolsep}{2.8pt}
    \resizebox{.88\linewidth}{!}{
		\begin{tabular}{p{1.8cm}<{\centering}|p{0.7cm}<{\centering}p{0.7cm}<{\centering}p{0.7cm}<{\centering}p{0.7cm}<{\centering}p{1.3cm}<{\centering}}
			\toprule
           $b$&100&200&400&1600&2048 (all) \\

			\midrule
			CT-{\it var} &60.6&\textbf{61.1}&60.8&59.0&58.3	\\
			CT-{\it var}+NCL&62.9&\textbf{63.4}&63.3&62.0&61.6	\\

			\bottomrule 
			
		\end{tabular}
		}
	
    \label{tab:b}
\end{table}

\begin{table}[t]

    \caption{Test accuracy (\%) on K200 with 80\% Symmetric Noise. Channels with higher score benefit more to model performance.}

	\centering
    \small
    \setlength{\tabcolsep}{3pt}
    \resizebox{.68\linewidth}{!}{
		\begin{tabular}{p{1.2cm}<{\centering}|p{1.2cm}<{\centering}p{1.2cm}<{\centering}p{1.2cm}<{\centering}}
			\toprule
            Score & Top & Middle & Bottom \\
			\midrule
			CT-{\it var}& \textbf{48.4} & 34.6  & 25.5        \\
            CT-{\it ave}& \textbf{48.1} & 35.4  & 20.9        \\
			\bottomrule 
		\end{tabular}
	}
	\label{tab:rank}

  \end{table}

\subsection{Regularization on Unlabelled Noisy Sample}
To fully utilize the unlabelled noisy set $\mathcal{D}_{\hat{u}}$, our Noise Contrastive Loss (NCL) provides a low-risk regularization in video classification scenario with noisy labels. We compare NCL with two famous contrastive losses: standard Contrastive Learning (CL) loss and Supervised Contrastive Learning (SCL)~\cite{khosla2020supervised} loss. CL loss involves all instances in training. As SCL requires the label supervisions, instances from $\mathcal{D}_{\hat{u}}$ are ignored in our SCL implementation. Training settings are shared for all methods, and the only difference is the contrastive strategy, which rules how positive/negative keys are selected. The results shown in Table~\ref{tab:k200} and Table~\ref{tab:k400} indicate that NCL can stably improve model performance.
We also implement three commonly used pseudo-label-based (PL) strategies, in which a pseudo label is assigned to each unlabelled noisy instance. PL-H~\cite{zhang2021flexmatch}/PL-S~\cite{li2019dividemix} generates a hard/soft label for unlabelled instances according to model prediction. PL-K~\cite{ortego2021multi} utilizes $k$-NN on the reference set $\Omega$ to predict pseudo label for instances in $\mathcal{D}_{\hat{u}}$. Here there are 64 nearest neighbors considered. We compare NCL with PL on the K200 dataset as shown in Table~\ref{tab:k200}. As we mentioned in Sec.\ref{subsec:ncl}, a pseudo label without robust post-processing is less reliable when the generalization of a model is weak. Tremendous wrong pseudo labels will dramatically degrade the supervision quality and thus hurt model prediction.



\subsection{Ablation Studies}
\label{sec:ablation}
\begin{table}[t]
    \caption{
    Comparison of Score Function on Different Datasets with 40\% Symmetric Noise (larger is better).}
	\centering
    \small
    \setlength{\tabcolsep}{4pt}
		\begin{tabular}{p{2.5cm}<{\centering}|p{1.2cm}<{\centering}p{1.2cm}<{\centering}}
			\toprule
            Score Function &   $\bm{\phi}_{ave}$&$\bm{\phi}_{var}$\\

			\midrule
			Mini-Kinetics &\textbf{94.9}&77.9\\
			Something V1 &45.5&\textbf{50.3}\\
			\bottomrule 
		\end{tabular}
	
	\label{Tab:intersection}
    \end{table}

\noindent\textbf{Parameter Sensitivity} The hyper-parameter $b$ controls how many dimensions are kept after channel truncation. As shown in Table~\ref{tab:b}, CT performs better when $b$ is close to 200. $b=2048$ means no channels is truncated.

\noindent\textbf{Channel Selection} We select the top, middle and bottom ranked $b$ channels, respectively, for noise detection and evaluate their impacts on final classification. As shown in Table~\ref{tab:rank}, model performance dramatically drops if we only keep $b$ channels with low scores, which proves the significance of channel selection.

\noindent\textbf{Score Function} We expect the proposed selections can keep as many the most discriminative channels for each category as the oracle selection does. To verify the ability of our proposed selections, we quantify the similarity of channel selection between the proposed selections and the oracle selection by the intersection size of selected channels by these two criteria at fifth epoch in model updating. The larger the size of the intersection, and the better the score function of the proposed selection. We report the evaluation in Table~\ref{Tab:intersection}. On the scene-based dataset K200, $\bm{\psi}_{ave}$ shares the most selected channels with the Oracle Selection $\bm{\psi}_{o}$. While on the temporal reasoning dataset SthV1, channels selected by the $\bm{\psi}_{var}$ is more discriminative. This observation keeps consistent with their performance in testing accuracy.

\noindent\textbf{Extension to Images} We also extend $\phi(\cdot)$ to an image version. Details can be found in Appendix.

\section{Discussion and Conclusion}
In this work, we propose an efficient framework for learning with noisy labels in video classification. This framework includes a feature-based noise detection method, the Channel Truncation, and a regularization approach, the Noise Contrastive Loss. The proposed CT adaptively selects discriminative channels for clean instances filtering in each model updating round, achieving good noise detection performance even with severe interference of noisy labels. The novel NCL learning strategy fully utilizes the detected noisy instances to assist model updating in reshaping the classification boundaries among categories. Extensive experiments validate the effectiveness of our framework NEAT in both noise detection and classification under various settings.

\section*{Acknowledgements}
This work was partially supported by the National Key R\&D Program of China (2022YFB4701400/4701402), SZSTC Grant (JCYJ20190809172201639, WDZC2020082020065-\\5001), and Shenzhen Key Laboratory (ZDSYS2021062309-\\2001004).

\bibliography{aaai23}

\clearpage

\newpage

\appendix

\section{Training Details}
When the proposed NEAT is evaluated on video datasets, we adapt all the training protocols from the work~\cite{lin2019tsm}.
The training hyper-parameters for all experiments on different datasets (the Kinetics 400, Mini-Kinetics, Something-Something-V1) are the same. Models are trained with 50 training epochs. The initial learning rate is 0.01 and it is decayed by 0.1 at epoch 20 and 40. The weight decay is set as $10^{-4}$. The batch size of training is 64, and the dropout is set as 0.5. The model is initialized from ImageNet pre-trained network. Videos from the Kinetics-400 and Mini-Kinetics are sampled by the I3D dense strategy. The uniform sampling strategy is applied on data from the Something-Something-V1 dataset during training phase. In the noise detection and testing phase, the uniform sampling strategy is utilized. Considering the efficiency, each instance consists of 8 frames in all the experiment settings. The projection head $h(\cdot)$ used in this paper is a single linear layer and the output dimension $\hat{d}$ of it is set as 128. The training protocols for all methods in the comparison experiment are kept the same.

For experiments on image datasets, similar to \cite{pleiss2020identifying}, a ResNet32 is trained from scratch. We choose SGD with a momentum of 0.9 and a weight decay of $5\times10^{-4}$ for optimization. The network is trained for 300 epochs. A small batch size of 8 is set to mitigate overfitting. The learning rate is initialized as $6.25\times10^{-3}$ and reduced by a factor of 10 at the 150-$th$ epochs. The length of warm up period is 10 epochs for CIFAR-10 and 30 epochs for CIFAR-100.

\begin{figure}[!t]
    \centering
    \begin{minipage}[b]{0.48\linewidth}
    \includegraphics[width=\textwidth]{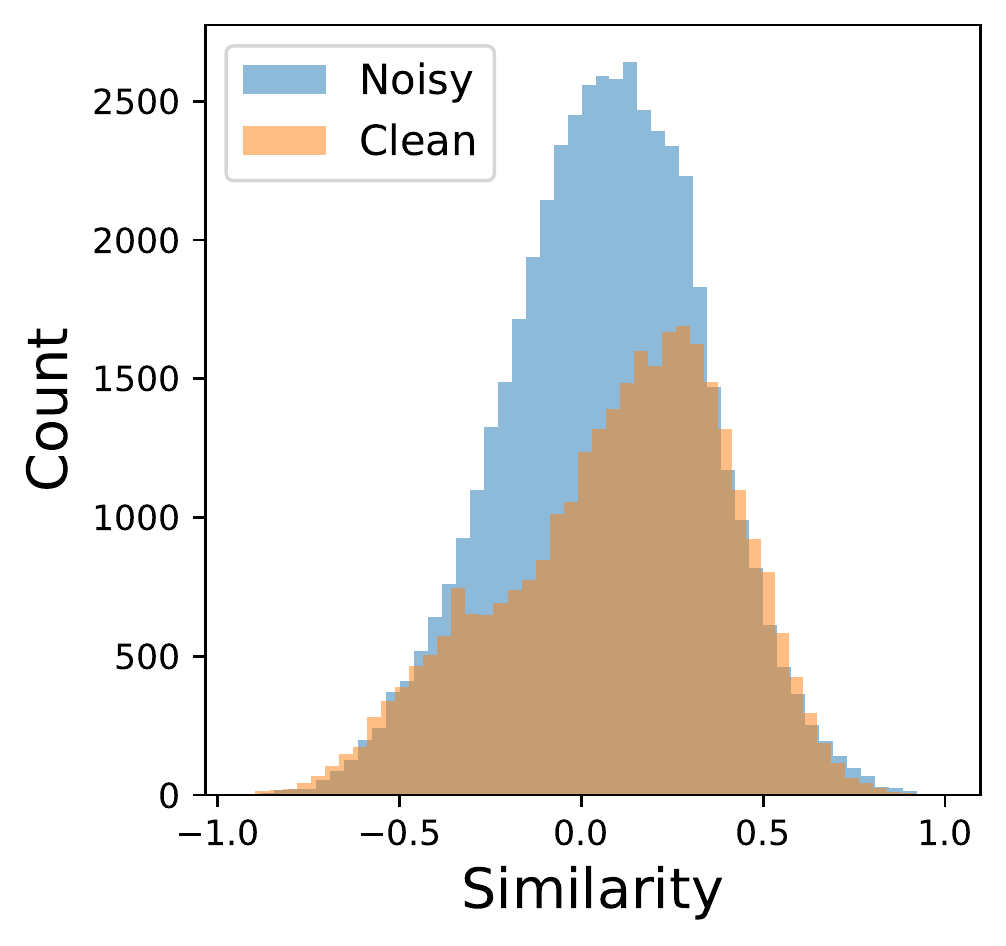}

    \end{minipage}
    \hfill
    \begin{minipage}[b]{0.48\linewidth}

    \includegraphics[width=\textwidth]{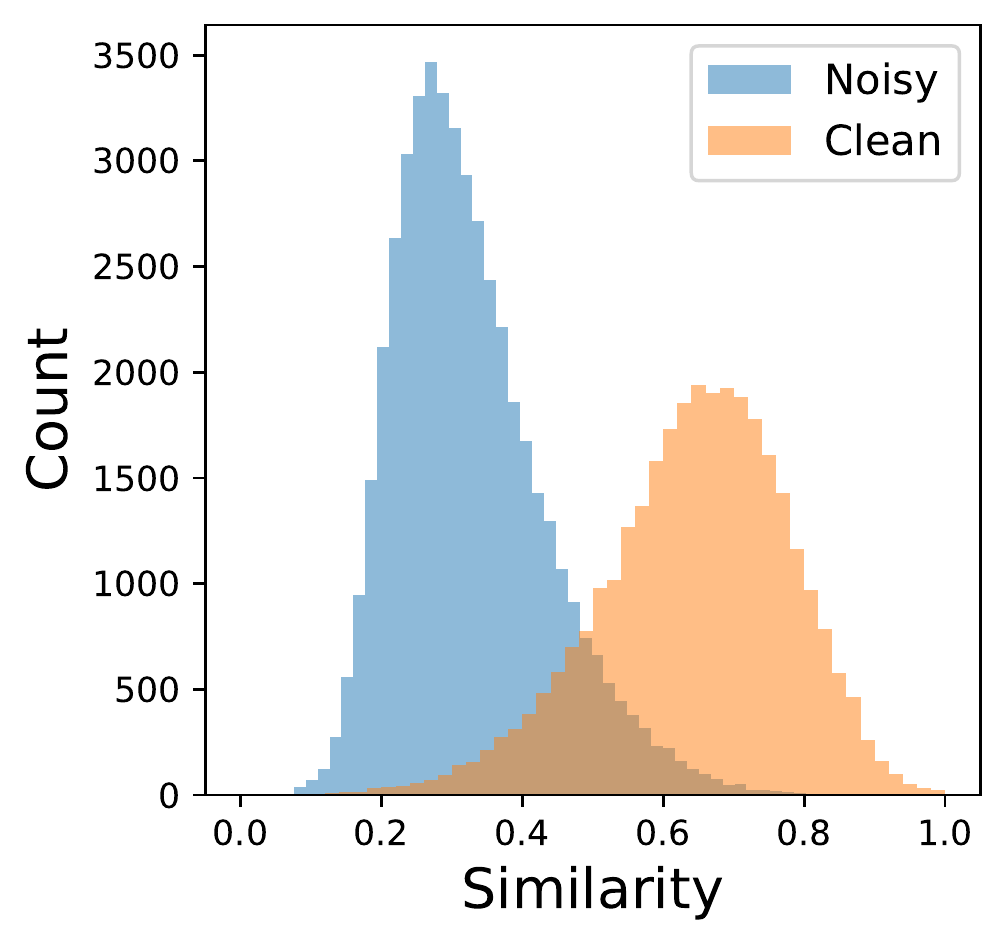}

    \end{minipage}

    \caption{\textbf{Left:} Similarity distribution of P-Correction. \textbf{Right:} Similarity distribution of CT-{\it var}. Both models are trained under 60\% close-set symmetric label noise on Mini-Kinetics dataset. This statistics comes from the fifth training epoch. All training protocols are kept the same.}
    \label{fig:pca}
\end{figure}

\section{Feature Selection v.s Linear Projection}

Principal Component Analysis (PCA) is a popular linear-projection-based dimension reduction technique, which projects the original feature into a subspace while maximizing the variance of new uncorrelated variables. By replacing Channel Truncation with PCA, we implement a PCA-based label noise detection method which we name as {\it P-Correction}. During the Noise Detection phase, P-Correction applies PCA on video-level features for each category individually and projects the original features into a $b$-dimensional subspace. The PCA projection matrix is updated in each noise detection round. After computing the similarities between each sample and the clean prototype in the subspace, P-Correction splits the original dataset into clean/noisy clusters by a GMM as our NEAT does.

\begin{table}[htbp]
    \caption{Testing Accuracies (\%) on Mini-Kinetics Dataset.}
    \label{tab:pcorrection}
	\begin{center}
    \small
    \setlength{\tabcolsep}{4pt}
		\begin{tabular}{l|cccc|ccc}
			\toprule
			\multicolumn{1}{c|}{Noise Type}&\multicolumn{4}{c}{Symmetric}&\multicolumn{3}{c}{Asymmetric} \\
			\multicolumn{1}{c|}{Noise Ratio}&20\%&40\%&60\%&80\%&10\%&20\%&40\%\\
			\midrule
            P-correction&66.8&46.1&53.2&30.9&68.7&67.9&57.0\\
            CT-{\it ave}&\textbf{69.4}&66.9&61.0&48.1&69.8&\textbf{68.6}&\textbf{58.4}\\
            CT-{\it var}&69.2&67.1&\textbf{61.1}&\textbf{48.4}&\textbf{70.0}&68.0&55.9\\
            \rowcolor{green!10} CT-{\it oracle} &70.4&67.7&61.5&49.6&70.6&70.0&58.9 \\
            \rowcolor[gray]{0.85} Clean Only &70.5&68.3&64.9&58.8&70.9&70.3&68.6 \\

			\bottomrule 
			
		\end{tabular}
	\end{center}
	\vspace{-10pt}
\end{table}

\begin{table}[t]
    \caption{Testing Accuracies (\%) on Mini-Kinetics Dataset with open-set label noise.}
    \label{tab:open}
	\begin{center}
    \small
    \setlength{\tabcolsep}{4pt}
		\begin{tabular}{l|p{1cm}<{\centering}p{1cm}<{\centering}|p{1cm}<{\centering}p{1cm}<{\centering}}
			\toprule
			Noise Type &\multicolumn{2}{c|}{K400-Symmetric}&\multicolumn{2}{c}{K400-Asymmetric} \\
			Noise Ratio &40\%&80\%&20\%&40\%\\
			\midrule
            Co-teaching\shortcite{han2018co}&61.2&21.7&59.9&54.6\\
            M-correction\shortcite{arazo2019unsupervised}&58.1&45.2&59.4&53.7\\
            CT-{\it var}&\textbf{68.2}&\textbf{48.4}&\textbf{69.9}&\textbf{63.9}\\
            \rowcolor[gray]{0.85} Clean Only &68.6&58.8&70.1&68.5 \\
            \midrule
            CT-{\it var} + NCL&\textbf{68.7}&\textbf{50.6}&\textbf{70.1}&\textbf{64.4}	 \\
            
			\bottomrule 
			
		\end{tabular}
	\end{center}
	
\end{table}

\begin{table}[tbp]
    \caption{Testing Accuracies (\%) on CIFAR-10 Datasets. "-" means diverged. The \textcolor{darkRed}{\textbf{best}} and \textcolor{darkBlue}{\textbf{second best}} results are marked. All results except DivideMix are from \cite{pleiss2020identifying}.}
    \vspace{-8pt}
    \label{tab:cifar10}
	\begin{center}
    \small
    \setlength{\tabcolsep}{4pt}
		\begin{tabular}{l|cccc|cc}
			\toprule
			\multicolumn{1}{c|}{Noise Type} &\multicolumn{4}{c}{Symmetric}&\multicolumn{2}{c}{Asymmetric} \\
			\multicolumn{1}{c|}{Noise Ratio} &20\%&40\%&60\%&80\%&20\%&40\%\\
			\midrule
            Bootstrap\shortcite{reed2015training}&77.6&62.6&48.0&31.2&76.2&55.0\\
            D2L\shortcite{ma2018dimensionality}&87.7&84.4&72.7&-&88.6&76.4\\
            $L_{\rm DMI}$\shortcite{xu2019l_dmi} &85.9&79.6&65.1&32.8&86.7&84.0\\
            Data Param\shortcite{saxena2019data}&82.1&70.8&49.3&18.9&82.1&55.5\\
            M-correction\shortcite{arazo2019unsupervised}&79.4&68.8&56.4&-&77.9&59.4\\
            INCV\shortcite{chen2019understanding}&89.5&86.8&81.1&53.3&88.3&79.8\\
            AUM\shortcite{pleiss2020identifying}&90.2&87.5&82.1&54.4&89.7&58.7\\
            DivideMix$^\dagger$\shortcite{li2019dividemix}&91.5&90.1&\textbf{\textcolor{darkBlue}{88.5}}&56.4&89.9&85.7\\
            DivideMix\shortcite{li2019dividemix}&\textbf{\textcolor{darkRed}{92.8}}&\textbf{\textcolor{darkRed}{91.4}}&\textbf{\textcolor{darkRed}{89.5}}&\textbf{\textcolor{darkRed}{78.8}}&\textbf{\textcolor{darkRed}{92.9}}&\textbf{\textcolor{darkBlue}{89.5}}\\
            \midrule
            CT-{\it img}&90.3&87.9&83.0&59.7&90.1&87.3\\
            CT-{\it img} + PL-H &87.2&86.5&83.5&62.0&87.3&79.9\\
            CT-{\it img} + PL-S &89.1&87.8&81.6&63.9&85.7&82.6\\
            CT-{\it img} + PL-K &89.6&88.0&84.8&68.5&87.9&83.4\\
            CT-{\it img} + NCL &\textbf{\textcolor{darkBlue}{91.9}}&\textbf{\textcolor{darkBlue}{90.3}}&85.8&\textbf{\textcolor{darkBlue}{69.8}}&\textbf{\textcolor{darkBlue}{92.0}}&\textbf{\textcolor{darkRed}{90.2}}\\
			\bottomrule 
			
		\end{tabular}
	\end{center}

\end{table}

In the experiments, we discover that the similarity distribution of clean and noisy instances in the PCA-subspace cannot form a two-peak shape, as shown in Figure~\ref{fig:pca}. This one-peak form leads to failure in GMM modeling, and therefore the clean and noisy instance cannot be well splitted. In this case, almost all samples are randomly alternately treated as clean or noisy ones during training. The test accuracy on Mini-Kinetics is shown in Table~\ref{tab:pcorrection} and our method (Channel Truncation) outperforms P-Correction under all noise settings. 

Besides, since there is no need to learn a linear projection matrix for each category, Channel Truncation is much faster than P-correction. In the comparison on Mini-Kinetics, Channel Truncation is on average around 12 times faster than P-Correction in noise detection.

\section{Open-set Noise on Mini-Kinetics}
The open-set noise comes from the set outside of the pre-defined label set. To evaluate the effectiveness of our methods, we conduct two different open-set label noises: 
\\[5pt]
\noindent\textbf{K400-Symmetric Noise}: Each noisy sample in the training set is randomly in-placed by a sample from categories that are in K400 but not in current dataset with a certain probability under uniform distribution.
\\[5pt]
\noindent\textbf{K400-Asymmetric Noise}: The samples in one category can only be replaced by the samples from a specific category that is in K400 but not in the current dataset. Such pair relationships do not overlap.

Table~\ref{tab:open} shows that CT-{\it var} outperforms the baseline methods under all open-set noise settings. 
Considering that the estimated noisy instance may not come from the pre-defined category, we limit the nearest neighbor set of the noisy query to the corresponding unlabeled set $\mathcal{D}_{\hat{u}}$. By this way, NCL can well handle the open-set noise without any further parameter tuning from the close-set noise setting.

\begin{table}[tbp]
    \caption{Testing Accuracies (\%) on CIFAR-100 Datasets. "-" means diverged. The \textcolor{darkRed}{\textbf{best}} and \textcolor{darkBlue}{\textbf{second best}} results are marked. All results except DivideMix are from \cite{pleiss2020identifying}.}
    \vspace{-8pt}
    \label{tab:image}
	\begin{center}
    \small
    \setlength{\tabcolsep}{4pt}
		\begin{tabular}{l|cccc|cc}
			\toprule
			\multicolumn{1}{c|}{Noise Type} &\multicolumn{4}{c}{Symmetric}&\multicolumn{2}{c}{Asymmetric} \\
			\multicolumn{1}{c|}{Noise Ratio} &20\%&40\%&60\%&80\%&20\%&40\%\\
			\midrule
            Bootstrap\shortcite{reed2015training}&51.4&41.1&29.7&10.2&53.4&38.7\\
            D2L\shortcite{ma2018dimensionality}&54.0&29.7&-&-&43.6&16.9\\
            $L_{\rm DMI}$\shortcite{xu2019l_dmi} &-&-&-&-&-&-\\
            Data Param\shortcite{saxena2019data}&56.3&46.1&32.8&11.9&56.2&39.0\\
            M-correction\shortcite{arazo2019unsupervised}&53.0&43.0&36.6&12.8&53.2&37.9\\
            INCV\shortcite{chen2019understanding}&58.6&55.4&43.7&23.7&56.8&44.4\\
            AUM\shortcite{pleiss2020identifying}&65.5&61.3&53.0&31.7&59.7&40.2\\
            DivideMix$^\dagger$\shortcite{li2019dividemix}&65.4&62.2&\textbf{\textcolor{darkBlue}{62.4}}&35.1&65.7&51.7\\
            DivideMix\shortcite{li2019dividemix}&67.6&\textbf{\textcolor{darkRed}{65.8}}&\textbf{\textcolor{darkRed}{63.7}}&\textbf{\textcolor{darkRed}{43.4}}&\textbf{\textcolor{darkBlue}{67.1}}&\textbf{\textcolor{darkBlue}{54.2}}\\
            \midrule
            CT-{\it img}        & 67.7 & 61.6 & 55.1 & 31.7 & 61.0 & 47.8 \\
            CT-{\it img} + PL-H & 67.4 & 62.1 & 53.6 & 27.2 & 65.0 & 51.1 \\
            CT-{\it img} + PL-S & \textbf{\textcolor{darkBlue}{68.2}} & 63.0 & 54.3 & 31.7 & 66.1 & 52.7 \\
            CT-{\it img} + PL-K & 67.7 & 62.7 & 53.6 & 31.7 & 66.6 & 53.5 \\
            CT-{\it img} + NCL  & \textbf{\textcolor{darkRed}{68.5}} & \textbf{\textcolor{darkBlue}{64.9}} & 57.5 & \textbf{\textcolor{darkBlue}{35.4}} &\textbf{\textcolor{darkRed}{67.4}}  & \textbf{\textcolor{darkRed}{58.5}} \\
			\bottomrule 
			
		\end{tabular}
	\end{center}
	\vspace{-10pt}
\end{table}

\section{Effectiveness on Image Datasets}

To further evaluate the effectiveness of NEAT, we conduct experiments on two popular image datasets CIFAR10 and CIFAR-100. To have a fair comparison with existing LNL methods on image classification, we follow the experiment settings in \cite{pleiss2020identifying}. 
ResNet-32 with random initialization is chosen as our backbone, which has 64 channels after the last layer.

A reasonable expansion of $\phi(\cdot)$ in image scenarios is:
$\phi_{img}(\bm{v}) = \bm{v}$, where $\bm{v}$ is the image representation. Channels with larger amplitudes are believed to have better discriminative ability. We keep $b=24$ channels after Channel Truncation in all experiments. We compare NEAT (CT and CT+NCL) with well-known methods for LNL on images, including Bootstrap\cite{reed2015training}, D2L\cite{ma2018dimensionality}, $L_{\rm DMI}$\cite{xu2019l_dmi}, Data Param\cite{saxena2019data}, INCV\cite{chen2019understanding}, DivideMix\cite{li2019dividemix} and AUM\cite{pleiss2020identifying}. We here re-implement two versions of DivideMix. One is the DivideMix with ResNet-32 as backbone. The other one (DivideMix$^\dagger$) is the same DivideMix without the Mixup~\cite{zhang2018mixup}, but with heavy post-processing of pseudo label kept. Note that Mixup~\cite{zhang2018mixup} help dramatically increase the size and diversity of training dataset, and thus there is a performance boost from DivideMix$^\dagger$ to DivideMix. While our proposed NEAT only takes the original dataset for training without any post-processing. As observed in Table~\ref{tab:cifar10}, and \ref{tab:image}, NEAT beats all other state-of-the-arts on both CIFAR-10 and CIFAR-100 datasets under several noise settings, and outperforms DivideMix, whose training data is diversified, by a large margin under severe noise setting Asymmetric-40\%. Our CT-{\it img} has beaten the best method, {\it i.e.} AUM, so far in noise detection. Besides, we also implement the image version of three PL strategies. As can be seen, pseudo label does help improve the classification performance on image data under several noise settings, but fails on videos. 
As before, NEAT achieves better classification performance than CT with the three PL strategies.




\section{Visualization}

\subsection{Channel Truncation Keeps Pivotal Channels}

In this section, we show heatmap visualizations to support our claim: the selected channels are closely related to the semantics, {\it e.g.} scene and motion, of a video clip in the specific category.

\begin{figure*}
    \centering
    \includegraphics[width=0.9\linewidth]{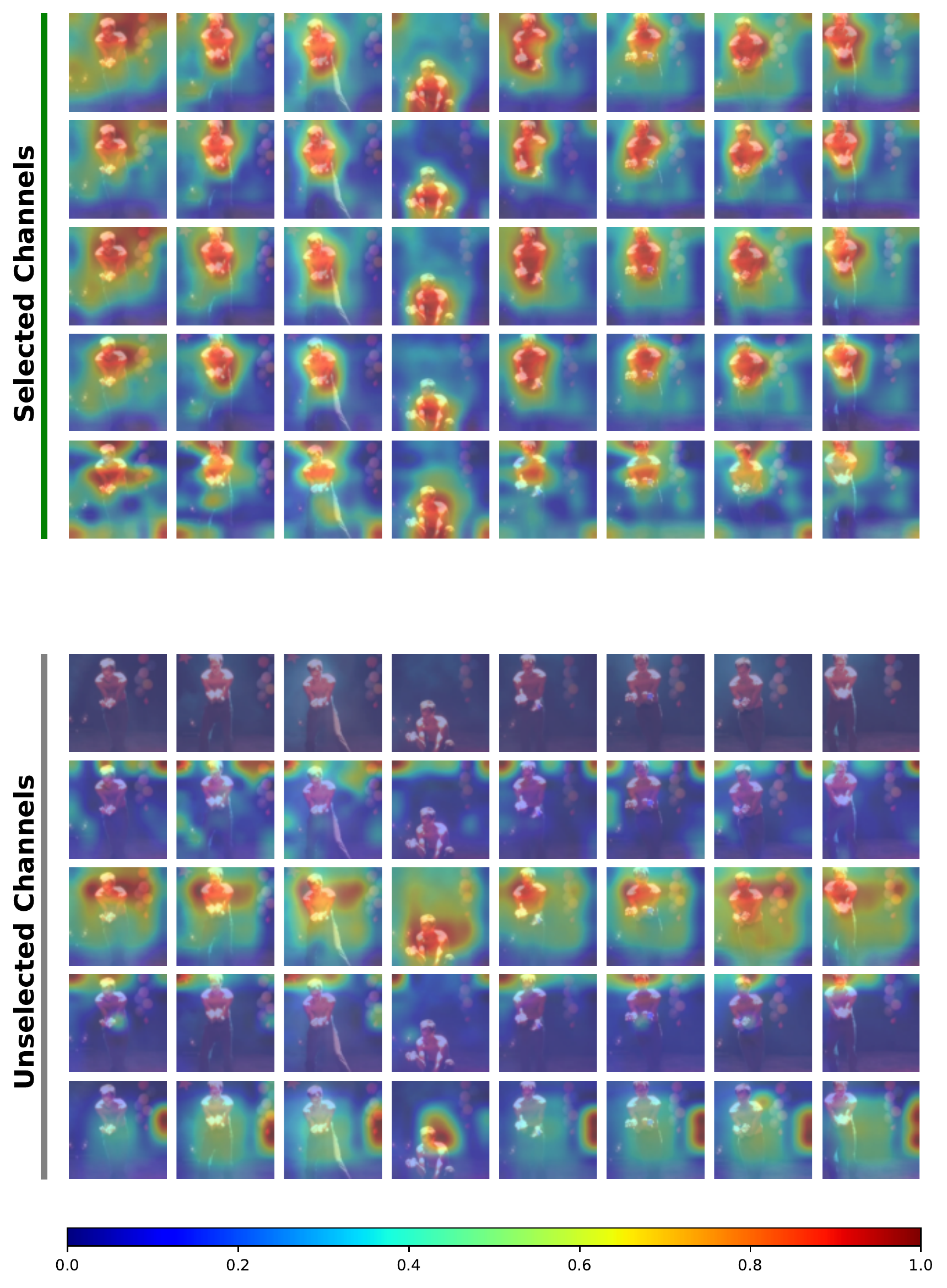}

    \caption{Visualization on a clip from {\it contact juggling}.}
    \label{fig:vis1}
\end{figure*}

\begin{figure*}
    \centering
    \includegraphics[width=0.9\linewidth]{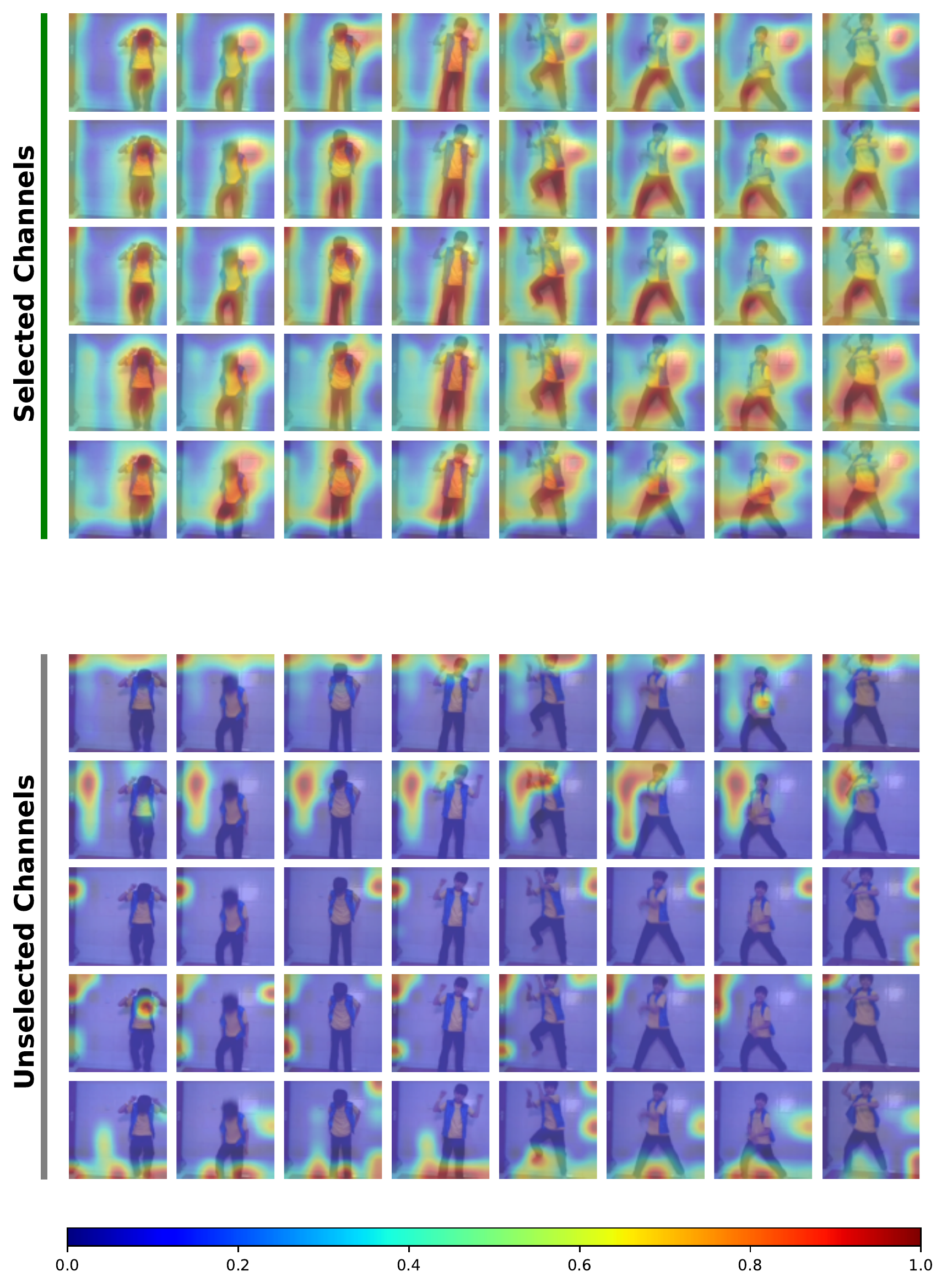}
    \caption{Visualization on a clip from {\it dancing gangnam style}.}
    \label{fig:vis2}
\end{figure*}

\begin{figure*}
    \centering
    \includegraphics[width=0.9\linewidth]{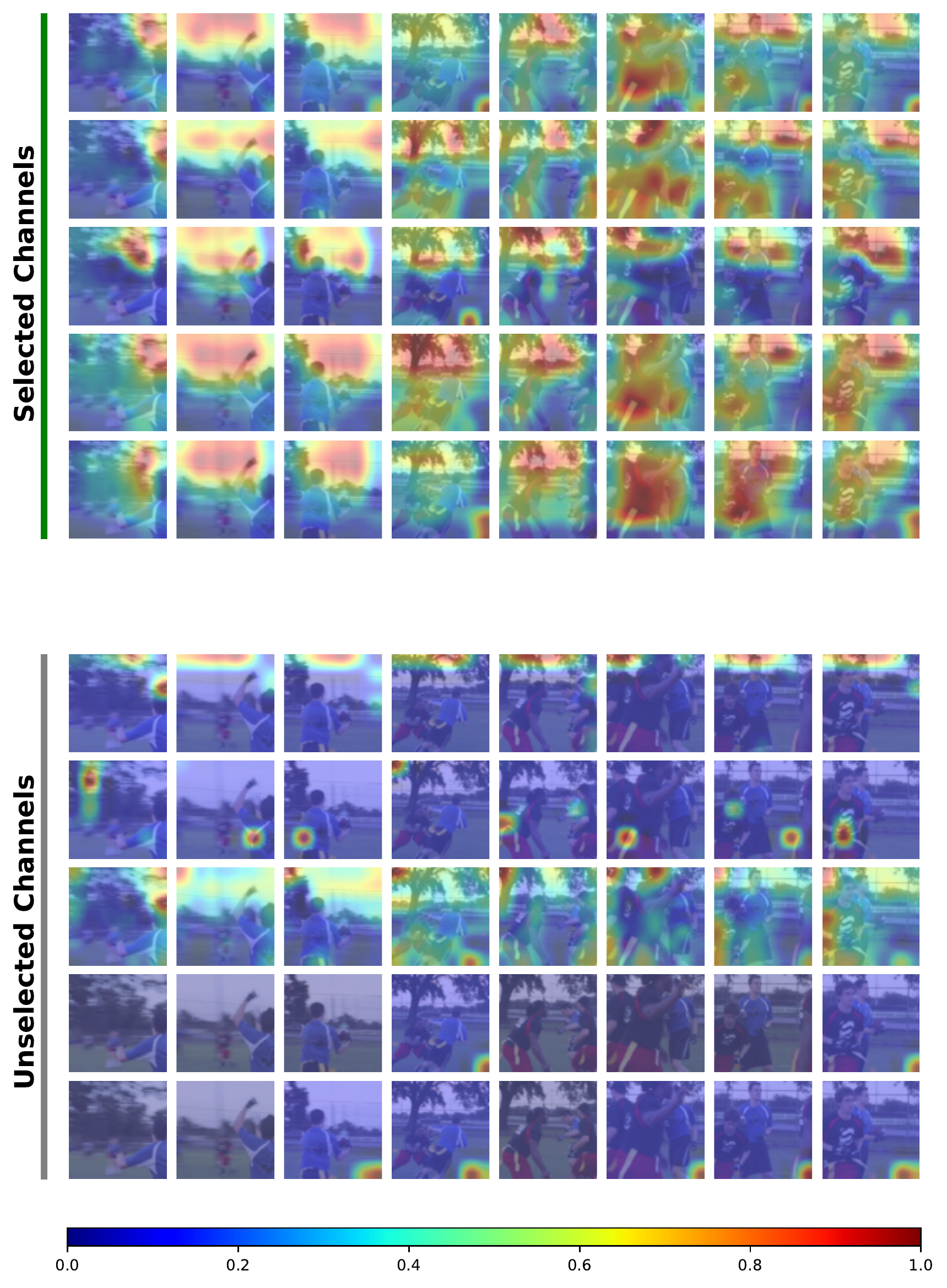}
    \caption{Visualization on a clip from {\it passing American football (in game)}.}
    \label{fig:vis3}
\end{figure*}

\begin{figure*}
    \centering
    \includegraphics[width=0.9\linewidth]{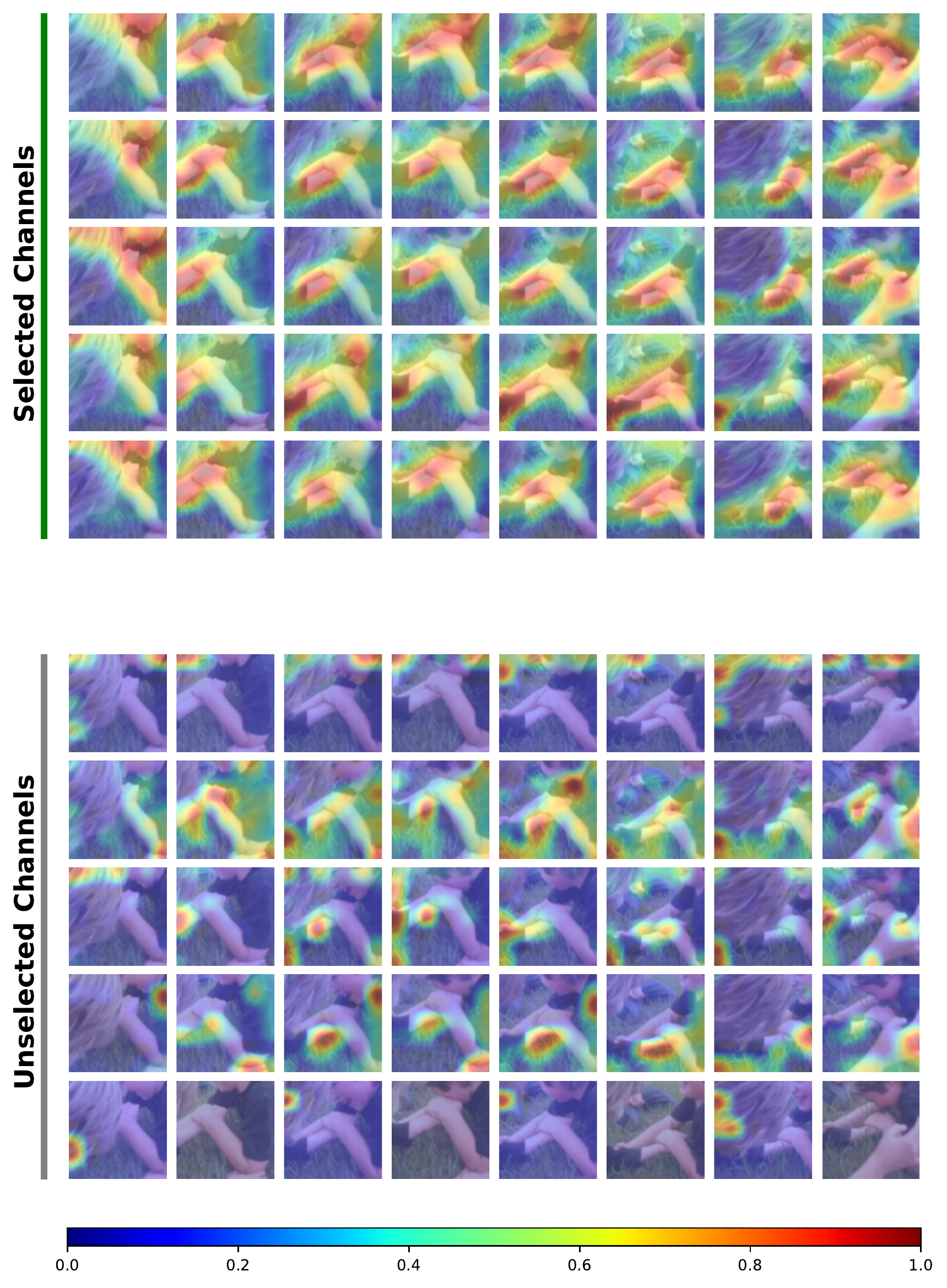}
    \caption{Visualization on a clip from {\it arm wrestling}.}
    \label{fig:vis4}
\end{figure*}

On Mini-Kinetrics, a network is trained for five epochs under 40\% symmetric noise setting. Then the frame-level feature map before the last pooling layer is visualized by using CAM introduced in \cite{zhou2016cvpr}. We here randomly show five heatmap sequences corresponding to the CT-selected channels and five truncated ones. The channel-relevant heatmap shows which region contributes more to the channel. The heatmaps are stacked with the original RGB frame for visualization in Figure~\ref{fig:vis1},~\ref{fig:vis2},~\ref{fig:vis3},~\ref{fig:vis4}. In our observation, channels selected by Channel Truncation tends to carry more semantic information that helps distinguish clean/noisy instances. 

\end{document}